\documentclass[nonacm,sigconf]{acmart}
\usepackage{caption} 

\AtBeginDocument{%
  \providecommand\BibTeX{{%
    \normalfont B\kern-0.5em{\scshape i\kern-0.25em b}\kern-0.8em\TeX}}}

\graphicspath{{./images/}} 
\usepackage{appendix}
\begin{document}

\title{Deciphering Air Travel Disruptions: A Machine Learning Approach}
\titlenote{GitHub Link: \href{https://github.com/jwgerlach00/flight_delay_prediction}{https://github.com/jwgerlach00/flight\_delay\_prediction}}

\author{Aravinda Jatavallabha}
\affiliation{%
  \institution{North Carolina State University}
  \city{Raleigh}
  \state{NC}
  \country{USA}
  \postcode{27695}}
\email{arjatava@ncsu.edu}

\author{Jacob Gerlach}
\affiliation{%
  \institution{North Carolina State University}
  \city{Raleigh}
  \state{NC}
  \country{USA}
  \postcode{27695}}
\email{jwgerlac@ncsu.edu}

\author{Aadithya Naresh}
\affiliation{%
  \institution{North Carolina State University}
  \city{Raleigh}
  \state{NC}
  \country{USA}
  \postcode{27695}}
\email{anaresh@ncsu.edu}

\begin{abstract}
This research investigates flight delay trends by examining factors such as departure time, airline, and airport. It employs regression machine learning methods to predict the contributions of various sources to delays. Time-series models, including LSTM, Hybrid LSTM, and Bi-LSTM, are compared with baseline regression models such as Multiple Regression, Decision Tree Regression, Random Forest Regression, and Neural Network. Despite considerable error in the baseline models, the study aims to identify influential features in delay prediction, potentially informing flight planning strategies. Unlike previous work, this research focuses on regression tasks and explores the use of time-series models for predicting flight delays. It offers insights into aviation operations by analyzing each delay component (e.g., security, weather) independently.
\end{abstract}


\maketitle

\section{Introduction}

Delay stands out as one of the most salient performance metrics for any transportation network. Flight punctuality is a critical aspect of airport and airline service quality, but flight delays in arrival and departure pose substantial challenges impacting operational efficiency and customer satisfaction \cite{1}. The Federal Aviation Administration (FAA) estimated in 2019 that these delays cost the aviation industry \$33 billion annually \cite{2}. In addition to financial implications, delays also contribute to environmental concerns through increased fuel emissions \cite{3,4}. Predicting flight delays is essential for proactive planning and resource allocation, benefiting airlines, passengers, and airports.\par

\begin{figure}[htp]
    \centering
    \includegraphics[width=5.5cm]{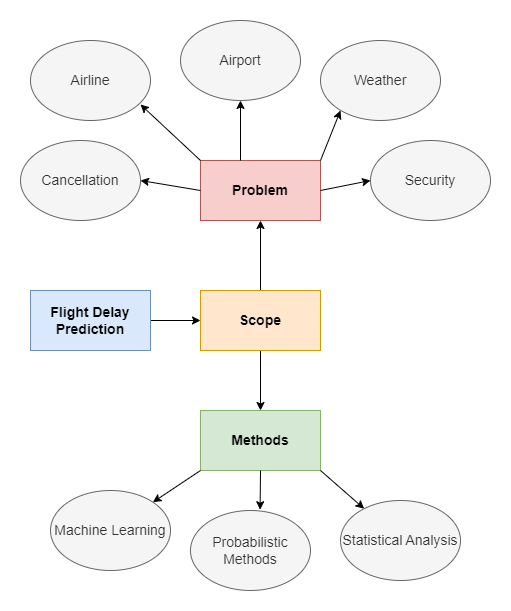}
    \caption{Taxonomy of Flight Delay Prediction Problem}
    \label{fig:tax}
\end{figure}

Flight delay prediction involves categorizing issues such as delay causes, institutional impacts, and mitigation strategies as seen in Fig \ref{fig:tax}. These encompass delay propagation, departure and route delays, and flight cancellations. While challenges persist, predictive tools aid operators and administrators in proactive management. Delays affect airlines, airports, and airspace, necessitating synchronized operations \cite{5}. Predictive system development entails utilizing methods like machine learning, probabilistic models, statistical analysis, or network representations.

With this study, we aim to individually predict the different factors contributing to flight delay rather than overall delay. We believe that doing so will allow for passengers and airlines to better predict flight delay bottlenecks.

\subsection{Statistical Analysis}

Utilizing statistical models involves employing correlation analysis, parametric and non-parametric tests, multivariate analysis, and econometric models. Government agencies have adopted these econometric models to comprehend the relationship between delays and factors such as passenger demand, fares, and aircraft size.

\subsection{Probabilistic Models}

Probabilistic modeling necessitates analysis tools that estimate the likelihood of an event based on historical data. The predicted outcome is expressed as a distribution function of probability. The inherent randomness factor invariably influences decisions or outcomes generated by probabilistic models.

\subsection{Machine Learning}

In supervised machine learning, datasets comprising input and output are provided, following which various algorithms are applied to analyze the data and establish mappings for new examples. \par

In this study, we have opted for a machine learning approach in our flight delay prediction methodology due to its inherent advantages over traditional statistical analysis and probabilistic modeling methods. Machine learning techniques, particularly supervised learning algorithms, offer unparalleled capabilities in handling complex and high-dimensional datasets commonly encountered in aviation data analysis. By automatically learning patterns and relationships from vast amounts of input data, these algorithms excel in capturing intricate dependencies between various factors influencing flight delays. Moreover, the flexibility and adaptability of machine learning models enable them to continuously update their predictions based on real-time data, ensuring timely adjustments to changing conditions and improving accuracy over time.

Furthermore, the versatility of machine learning algorithms in incorporating diverse sources of information, such as weather patterns, flight schedules, and aircraft characteristics, enhances the robustness of the predictive model. Unlike traditional statistical methods, which may struggle to adequately capture nonlinear relationships and interactions in complex data, machine learning models offer superior generalization capabilities, making them well-suited for real-time prediction tasks in dynamic environments like the aviation industry. Thus, by leveraging the power of machine learning, we aim to develop a highly accurate and adaptable flight delay prediction system capable of addressing the complexities and uncertainties inherent in aviation operations.

\section{Literature Survey}
 Several studies have been done in the field of flight delay prediction. One of the research \cite{6} evaluates machine learning models for airline flight delay prediction using JFK airport data, achieving a remarkable accuracy of 97.78\% with the Decision Tree model. Random Forest and Gradient Boosted Tree follow closely, showcasing accuracies of 92.40\% and 93.34\%, respectively. Future investigations suggested focusing on refining data imbalance and exploring alternative ensemble techniques for heightened prediction precision.
 
 In another study \cite{7}, the researchers introduced a method using the Support Vector Regressor (SVR) algorithm to predict flight delays at U.S. airports. To handle the massive amount of data, they organized and
 sampled it month by month. They initially used cat-boost, a method that boosts categorical variables, to evaluate each feature’s importance, ultimately selecting 15 key features to train the model. Following this, regression models were used to predict the specific delay time.
 
 In a different research study \cite{8}, the authors employed a Support Vector Machine (SVM) to delve into the factors influencing air traffic delays at three key New York City airports. They examined various explanatory variables to uncover their relationship with flight delays, airport operations, and flow management. By calculating the probabilities of these variables contributing to delays and comparing them, they gained deeper insights into the causes of departure delays.
 
 The proposed CNN-LSTM deep learning framework by the authors \cite{9} offers a novel approach to address
 the complexities of flight delay prediction, crucial for optimizing flight scheduling, airline operations, and airport management. Leveraging Convolutional Neural Network (CNN) and Long Short-Term Memory (LSTM) architectures, their model captures both spatial and temporal correlations inherent in flight delays. By integrating extrinsic features and employing a random forest model, they achieved an impressive prediction accuracy of 92.39\% on U.S. domestic flights in 2019. This study underscores the potential of our approach in aiding airport regulators to proactively manage delays and enhance overall airport performance.

In another study \cite{10}, the authors performed a classification task utilizing machine learning algorithms like Random Forest, Multi-Layer Perceptron, Naive Bayes Classifier, and KNN. The performance of the Random Forest algorithm surpassed other algorithms used achieving an accuracy of 83\%.

\section{DATASET DESCRIPTION AND Preprocessing}
\subsection{Description}
The dataset used in this study is obtained from the U.S. Department of Transportation, Bureau of Transportation Statistics from January 2019 – August 2023 \cite{11}. It contains 32 attributes related to planned flight date-time, airline, planned origin and destination, cancellation and diversion status, overall delay, and delay due to individual components (carrier, weather, NAS, security, late aircraft), among others. Table.\ref{tab:dataset_attributes} includes the full list of attributes along with their descriptions.

Our aim is to use this data to build regression models that can predict delay due to each component. We benchmark this model with models built to predict overall arrival delay.

\textbf{Hypothesis:} We believe that delay component prediction will perform just as well on overall delay prediction as just predicting arrival delay while offering increased explainability. We also believe that a time-series architecture will outperform baseline models.

\subsection{Pruning Data}
In our dataset, 2.87\% of records resulted in a canceled or diverted flight. These data entries were removed as they are beyond the scope of predicting flight delay (minutes). An additional 79.3\% of the data entries have missing values for all five individual delay components. These entries were removed as they provide inadequate information for predicting flight delays due to delay components.

It was verified that for all remaining records, the sum of delay components ("DELAY\_DUE\_CARRIER", "DELAY\_DUE\_WEATHER", \\"DELAY\_DUE\_NAS", "DELAY\_DUE\_SECURITY", and \\"DELAY\_DUE\_LATE\_AIRCRAFT") is equal to the overall arrival delay ("ARR\_DELAY"). 

\begin{figure}[htp]
    \centering
    \includegraphics[width=9cm]{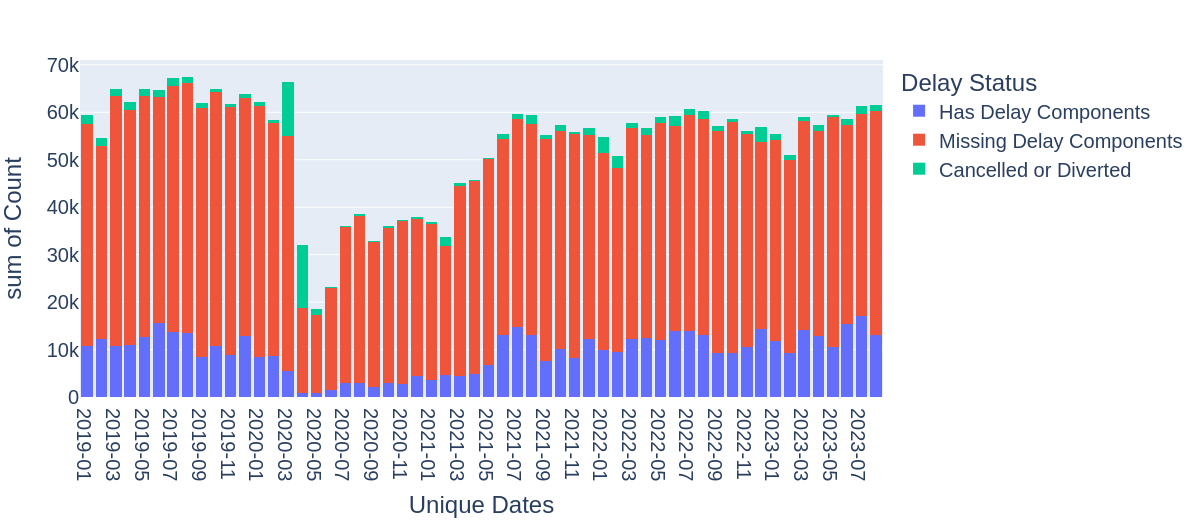}
    \caption{Flight record counts}
    \label{fig:flight_counts}
\end{figure}

The distribution of the data concerning time is depicted in Fig. \ref{fig:flight_counts}, with blue indicating the retained data containing delay components. Despite the removal of the majority of the data, 533,863 flight records with a reasonably consistent temporal distribution remain.

To identify outliers, we used the interquartile range (IQR) method, a widely accepted technique for outlier detection and removal. Entries with overall arrival delay ("ARR\_DELAY") outside of the bounds Q1 - (1.5 * IQR) and Q3 + (1.5 * IQR) were removed. This resulted in the removal of 8.248\% of remaining entries, or 44,035 flight records. All removed flight records have an arrival delay of over 154 minutes, and no flight record was below the lower bound. What remains are 489,828 flight records with the arrival delay properties in Table.\ref{tab:outlier_removal}. Standard deviation is notably reduced which can aid in improved model generalization and stability with the caveat that our model will not be valid for accurately predicting delay over 154 minutes.

\begin{table}[htp]
    \centering
    \small
    \captionsetup{justification=centering}
    \caption{Statistics of Arrival Time (minutes) Before and After Outlier Removal} 
    \begin{tabular}{|c|c|c|}

    \hline
     & \textbf{Pre-Outlier Removal} & \textbf{Post-Outlier Removal} \\
    \hline
    Mean & 67.526 & 47.828 \\
    Standard Dev. & 93.909 & 32.869 \\
    Minimum & 15 & 15 \\
    Maximum & 2934 & 154 \\
    \hline 
    \end{tabular}

    \label{tab:outlier_removal}
\end{table}

Pearson's Correlation was used to compute the correlation between the continuous independent attributes: "CRS\_DEP\_TIME", "TAXI\_OUT", "CRS\_ARR\_TIME", "TAXI\_IN", "CRS\_ELAPSED\_TIME", "DISTANCE" and dependent attribute "ARR\_DELAY". Table.\ref{tab:correlation} shows the results. We decided to include all of these attributes except for "CRS\_ELAPSED\_TIME" as it is a result of "CRS\_ARR\_TIME" and "CRS\_DEP\_TIME". PCA was attempted but did not result in strengthened correlations.

Pearson's Correlation is not valid for categorical attributes. For the categorical independent attributes we eliminated "AIRLINE\_DOT", "DOT\_CODE", "AIRLINE\_CODE", "DOT\_CODE" as they are redundant with "AIRLINE". We also eliminated "ORIGIN\_CITY" as it is redundant with "ORIGIN", and "DEST\_CITY" as it is redundant with "DEST". These redundancies were proven using The Kruskal-Wallis H-test.

Our feature set consists of these included continuous and categorical independent attributes. We constructed on label-set, consisting of delay due to 5 components. Table.\ref{tab:variables} shows this break-down. We also constructed a label set consisting of just the overall arrival delay, "ARR\_DELAY".

\begin{table}[htbp]
\centering
\captionsetup{justification=centering}
\caption{Pearson's Correlation of Each Attribute concerning Arrival Delay} 
\small
\begin{tabular}{|c|c|}
\hline
\textbf{Attribute} & \textbf{Pearson's Correlation} \\
\hline
CRS\_DEP\_TIME & 0.0704 \\
TAXI\_OUT & 0.0541 \\
CRS\_ARR\_TIME & 0.0500 \\
TAXI\_IN & 0.0235 \\
CRS\_ELAPSED\_TIME & -0.0122 \\
DISTANCE & -0.0228 \\
\hline 
\end{tabular}

\label{tab:correlation}
\end{table}

\subsection{Data Transformation}
All time values in the form of "hhmm" (Table.\ref{tab:dataset_attributes}) were converted to minutes.

For label-encoding, "FL\_DATE" in the form of YYYY-MM-DD was transformed into 3 separate attributes: YEAR, MONTH, and DAY. All categorical attributes were encoded into integer labels using sklearn.preprocessing.LabelEncoder to prepare the dataset for modeling. One-hot encoding the data would massively increase the feature size to a degree which was unrealistic given the computing power we had access to.

The dataset was split chronologically into training and testing sets for all models. Specifically, a 75\%/25\% split was performed, with 75\% of the data allocated to the training set and 25\% to the testing set. This chronological splitting ensured that both training and testing sets preserved the temporal order of the data, allowing all models to effectively utilize time series information for prediction.

\begin{table}[htbp]
\centering
\small
\captionsetup{justification=centering}
\caption{Pearson's Correlation of Each Attribute Concerning Arrival Delay} 
\begin{tabular}{|c|c|c|}
\hline
\textbf{Independent Vars.} & \textbf{Dependent Delay Components} \\
\hline
CRS\_DEP\_TIME & DELAY\_DUE\_CARRIER \\
TAXI\_OUT & DELAY\_DUE\_WEATHER \\
CRS\_ARR\_TIME & DELAY\_DUE\_SECURITY \\
TAXI\_IN & DELAY\_DUE\_NAS \\
DISTANCE & DELAY\_DUE\_LATE\_AIRCRAFT \\
YEAR & \\
MONTH & \\
DAY & \\
AIRLINE & \\
ORIGIN & \\
DEST & \\
\hline 
\end{tabular}

\label{tab:variables}
\end{table}

\section{PROPOSED METHOD}
This study makes a careful selection of time series forecasting models known for their adeptness in capturing intricate temporal dependencies and patterns. Our choices include LSTM, Bi-LSTM, and a hybrid architecture LSTM+CNN. These models are chosen precisely for their capacity to handle the dynamic nature of flight delay data, where temporal relationships play a crucial role. Additionally, to establish a benchmark for comparison, a set of baseline regression models has also been included. These models, comprising Multiple Regression, Random Forest Regression, Decision Tree Regression, XGBoost, and Artificial Neural Network, offer a traditional yet robust approach to forecasting flight delays. By juxtaposing the performance of advanced time series models against these baseline regressors, we aim to elucidate the efficacy and superiority of our chosen methodologies in predicting flight delays accurately.\par

 LSTM and Bi-LSTM models will be particularly advantageous due to their ability to handle vanishing gradient problems and capture long-term dependencies inherent in time series data. Techniques such as gradient clipping and batch normalization have been employed to mitigate the vanishing gradient issue in LSTM models. The training process involves iterating over the dataset multiple times (epochs) and adjusting the model's weights using optimization algorithms.\par

The anticipation is that the time series forecasting models, particularly LSTM and its variants, will outperform the baseline regression models due to their ability to capture temporal dependencies inherent in flight delay data. By comparing the performance of different models and analyzing their strengths and weaknesses, we aim to provide valuable insights into the effectiveness of various machine learning approaches for flight delay prediction. \par

\section{EXPERIMENT}

In this section, we discuss the various models utilized in our study for predicting flight delays. We present a detailed overview of each model, including LSTM, Bi-LSTM, and hybrid architecture LSTM+CNN, chosen for their capability to capture temporal dependencies and patterns within flight delay data. Additionally, we describe the baseline regression models, such as Multiple Regression, Decision Tree Regression, Random Forest Regression, XGBoost, and Artificial Neural Network, employed for comparative analysis.

\subsection{Multiple Regression}
In our case of flight delay prediction, we have utilized the Multiple Regression model primarily for its simplicity and as a baseline for comparing with other more complex models. Multiple Regression is a simple yet powerful technique used for predicting continuous outcomes. It assumes a linear relationship between the independent variables \( X \) and the dependent variable \( Y \), represented by the equation:

\[
Y = \beta_0 + \beta_1X_1 + \beta_2X_2 + ... + \beta_nX_n + \epsilon
\]

where:
\begin{itemize}
  \item \( Y \) is the dependent variable,
  \item \( X_1, X_2, ..., X_n \) are the independent variables,
  \item \( \beta_0, \beta_1, ..., \beta_n \) are the coefficients, and
  \item \( \epsilon \) is the error term.
\end{itemize}

The model aims to minimize the sum of squared residuals between the observed and predicted values of \( Y \), thus estimating the coefficients \( \beta_0, \beta_1, ..., \beta_n \) that best fit the data.

Using the independent and dependent variables mentioned in Table.\ref{tab:variables}, the model is trained.

\subsection{Decision Tree Regressor}
\begin{figure}[h]
    \centering
    \captionsetup{justification=centering}
    \includegraphics[width=0.4\textwidth]{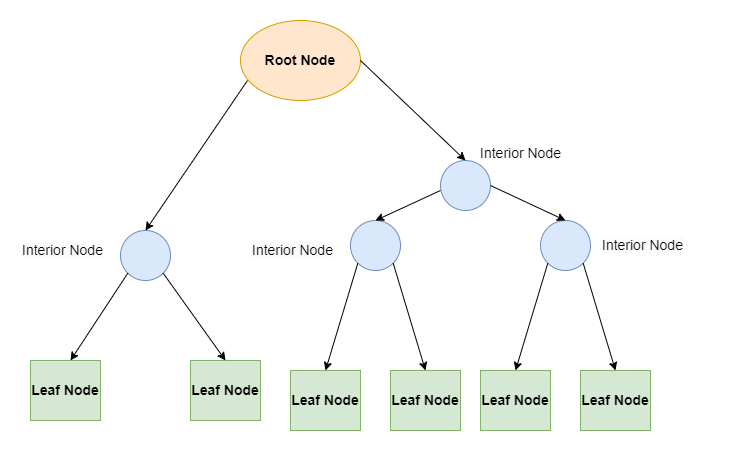}
    \caption{Structure of Decision Tree Nodes: Root, Interior, and Leaf.}
    \label{decisiontree}
\end{figure}

A non-parametric supervised learning model known as Decision Trees (DT) is utilized for both classification and regression tasks. Decision trees, organized in a tree-like structure, consist of root, interior, and leaf nodes representing different aspects of the dataset and decision-making procedures. This method proves beneficial for decision-oriented problems. Fig. \ref{decisiontree} illustrates the structure of the decision tree.

In our study, we selected the Decision Tree regressor as a baseline model for flight delay prediction, given its simplicity and interpretability, making it suitable for comparison with more complex models. Additionally, decision trees are versatile in handling both numerical and categorical data, thus accommodating the diverse features present in flight delay datasets. However, Decision Tree regressors may not perform optimally for flight delay prediction, especially with time-stamped data, due to several limitations. These include a tendency to overfit training data, particularly in high-dimensional datasets with intricate temporal dependencies, leading to poor generalization performance on unseen data. Moreover, decision trees may struggle to capture complex nonlinear relationships between features and flight delays in dynamic aviation environments. Furthermore, their reliance on binary splits based on certain feature thresholds may not effectively capture the nuanced patterns present in flight delay data, resulting in suboptimal decision boundaries. Thus, while Decision Tree regressors offer simplicity and interpretability, they may not be well-suited for accurately predicting flight delays, particularly in the context of time-stamped data with complex temporal dynamics. Therefore, more sophisticated models such as LSTM and hybrid architectures are necessary to achieve better predictive performance in this domain.\par

In our study, we have chosen a maximum depth of 10 and a minimum number of sample leaves of 5 to train our model.\par

\subsection{Random Forest Regressor}
Random Forest model was selected as a baseline model due to its simplicity, versatility, and robustness \cite{17}. Random Forest employs an ensemble of decision trees, known as a forest. Each decision tree is constructed on randomly selected data points from the dataset, and predictions from all trees are aggregated to predict the output variable. Unlike a single decision tree, Random Forest efficiently handles missing values and mitigates overfitting by averaging the results across multiple trees. The architecture of a random forest model is also illustrated by Fig. \ref{fig:RF} in the appendix.

Despite its advantages, Random Forest Regressors face challenges when applied to time-stamped data, such as that in our study. Time-stamped data inherently contains temporal dependencies and sequential patterns that Random Forests struggle to capture effectively. These models treat each data point independently, disregarding the sequential nature of observations, which is crucial for accurate prediction of flight delays. Moreover, while Random Forests are less prone to overfitting compared to single decision trees, they may still struggle with extrapolation beyond the training data, particularly when the temporal relationships between variables are not adequately captured.
\sloppy
In our study, we employed a sklearn.ensemble.RandomForestRegressor using a maximum depth of 20 and 100 for the number of estimators.


\subsection{XGBoost Regressor}
The extreme gradient boosting (XGBoost) algorithm \cite{12} is a powerful tree-based ensemble learning method, widely used in data science research. It builds on gradient-boosting architecture, utilizing various complementary functions to estimate results. XGBoost's objective function, as shown in Eq. \ref{eq:prediction}, predicts the output $\bar{y}_i$ for the $i$th data point \cite{13}\cite{14}.

\begin{equation}
\bar{y}_i = y_{0i} + \eta \sum_{k=1}^{n} f_k(U_i) \label{eq:prediction}
\end{equation}

Other equations involved in the calculations of the algorithm and the model architecture illustrated by Fig. \ref{fig:xgboost} are included in the appendix section.

XGBoost offers advantages such as robustness, flexibility in hyperparameter tuning, and efficient handling of large datasets. However, when applied to time series data like flight delay prediction, it may face challenges due to its static feature reliance and potential difficulties in capturing dynamic temporal dependencies.

Despite its advantages, XGBoost's suitability for time series data depends on effective feature engineering and model tuning. In the context of flight delay prediction, where temporal dynamics play a crucial role, XGBoost may not fully capture complex relationships, leading to suboptimal performance compared to specialized time series forecasting techniques.

\subsection{Artificial Neural Network}
An artificial neural network (ANN) is a computational system designed to emulate the structure and function of the human brain's cerebral cortex. Comprising interconnected processing units, ANNs mimic the brain's ability to process information and learn from it. In the brain, neurons play a central role by gathering sensory data through dendrites, processing it, and transmitting it via axons. Similarly, in an artificial neural network, an input layer receives input data through multiple neurons, while an output layer sends processed output to external systems. Hidden layers, situated between the input and output layers, perform complex transformations on the input data. A simple architecture of a neural network model illustrated by Fig. \ref{fig:ann} can be found in the appendix section.
In our model, a neural network architecture was implemented using TensorFlow in Python, comprising an input layer followed by two hidden layers. The first hidden layer consists of 64 neurons with a ReLU activation function, while the second has 32 neurons with the same activation function. The output layer matches the number of columns in the output variable Y, containing 5 neurons. We trained the model with the Adam optimizer, Mean Squared Error (MSE) loss function, and Mean Absolute Error (MAE) as the evaluation metric. The model underwent 50 epochs of training with a batch size of 32 and 20\% validation split.

\subsection{Long Short-Term Memory (LSTM)}
\label{lstm section}
\begin{figure}[h]
    \centering
    \captionsetup{justification=centering}
    \includegraphics[width=0.4\textwidth]{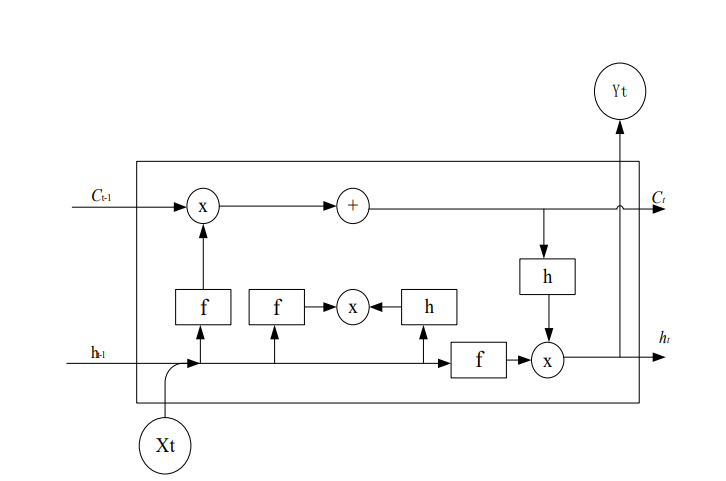}
    \caption{LSTM Unit Structure}
    \label{lstm-arc}
\end{figure}

Long Short-Term Memory (LSTM) models provide a powerful framework for modeling sequential data, such as time series prediction tasks like flight delay prediction. LSTMs are a type of recurrent neural network (RNN) architecture designed to address the vanishing gradient problem and capture long-term dependencies in sequential data effectively. The architecture of an LSTM cell is illustrated by Fig.\ref{lstm-arc}.

The core components of an LSTM unit include a cell state ($C_t$), an input gate ($i_t$), a forget gate ($f_t$), an output gate ($o_t$), and the hidden state ($h_t$). These components are governed by the following equations:

\[
\begin{aligned}
i_t &= \sigma(W_{xi} x_t + W_{hi} h_{t-1} + b_i) \\
f_t &= \sigma(W_{xf} x_t + W_{hf} h_{t-1} + b_f) \\
o_t &= \sigma(W_{xo} x_t + W_{ho} h_{t-1} + b_o) \\
g_t &= \tanh(W_{xg} x_t + W_{hg} h_{t-1} + b_g) \\
C_t &= f_t \odot C_{t-1} + i_t \odot g_t \\
h_t &= o_t \odot \tanh(C_t)
\end{aligned}
\]

Where:
\begin{itemize}
  \item $x_t$ is the input vector at time step $t$.
  \item $h_{t-1}$ is the hidden state vector from the previous time step.
  \item $W$ and $b$ are weight matrices and bias vectors that are learned during training.
  \item $\sigma$ is the sigmoid activation function, and $\odot$ represents element-wise multiplication.
  \item $i_t$, $f_t$, $o_t$, and $g_t$ denote the input gate, forget gate, output gate, and cell gate activations, respectively.
\end{itemize}

During training, the LSTM model learns to update its cell state ($C_t$) and hidden state ($h_t$) based on the input at each time step, allowing it to capture temporal dependencies and make predictions.

In our study of flight delay prediction, 
The Long Short-Term Memory (LSTM) model was selected for adeptness in handling time series data, particularly in capturing intricate temporal dependencies present in historical flight data for predicting flight delays \cite{16}. Unlike conventional recurrent neural networks (RNNs), LSTMs are equipped with mechanisms to retain information over extended sequences, mitigating the challenge of vanishing gradients and allowing for more effective learning of long-range dependencies inherent in time series data. In our study of flight delay prediction, this capability is crucial as flight delays are influenced by a multitude of factors such as departure times, weather conditions, and airline schedules, all of which exhibit complex temporal patterns.\par

The LSTM architecture's inherent ability to retain and selectively update information over time enables the model to effectively capture both short-term fluctuations and long-term trends in the data. By leveraging this feature, our LSTM-based approach aims to extract meaningful insights from historical flight data, enabling more accurate predictions of flight delays. Moreover, the flexibility of LSTM models in handling multivariate time series data allows us to incorporate diverse features such as historical delays, airport congestion levels, and flight routes, providing a comprehensive understanding of the factors contributing to flight delays. 

\begin{figure}[h]
    \centering
    \captionsetup{justification=centering}
    \includegraphics[width=0.25\textwidth]{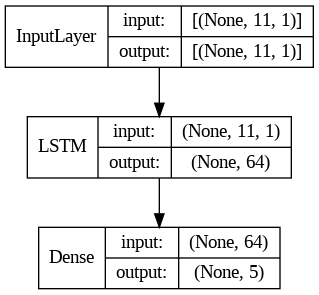}
    \caption{LSTM architecture}
    \label{lstm}
\end{figure}

The architecture of the model used in the study is illustrated in Fig \ref{lstm}.

\subsection{Bidirectional Long Short-Term Memory (BiLSTM)}\label{bilstm section}
Bidirectional Long Short-Term Memory (BiLSTM) models extend the capabilities of LSTM by processing input sequences in both forward and backward directions. This allows the model to capture contextual information from both past and future time steps, making it particularly effective for tasks requiring a comprehensive understanding of sequence data.

\begin{figure}[h]
    \centering
    \captionsetup{justification=centering}
    \includegraphics[width=0.4\textwidth]{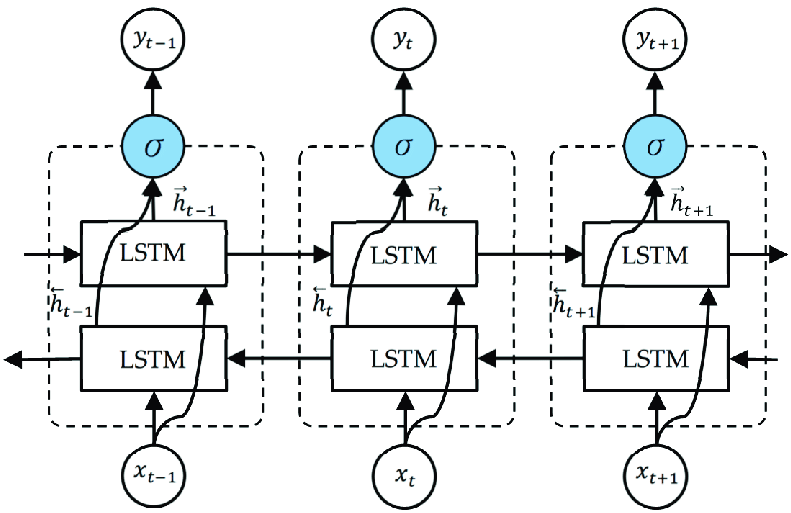}
    \caption{The unfolded architecture of Bidirectional LSTM (BiLSTM) with three consecutive steps \cite{17}}
    \label{bilstm}
\end{figure}

The architecture of a BiLSTM unit is similar to that of LSTM, with the addition of a backward LSTM layer which is illustrated by Fig.\ref{bilstm}. At each time step \(t\), the input sequence is fed into both the forward (\(\rightarrow\)) and backward (\(\leftarrow\)) LSTM layers, producing two separate hidden states \(h_t^{(\rightarrow)}\) and \(h_t^{(\leftarrow)}\) for the forward and backward directions, respectively.

The output of the BiLSTM layer at each time step is typically a concatenation of the forward and backward hidden states:

\[
h_t = [h_t^{(\rightarrow)}, h_t^{(\leftarrow)}]
\]

This allows the model to capture dependencies from both past and future contexts, enabling it to make more informed predictions based on the entire sequence.

In our study of flight delay prediction, we leverage the capabilities of BiLSTM models to capture bidirectional dependencies in historical flight data. By processing input sequences in both forward and backward directions, BiLSTM models can effectively capture temporal patterns and contextual information from past and future time steps, leading to more accurate predictions of flight delays.

The flexibility and effectiveness of BiLSTM models make them a valuable tool for time series analysis tasks where capturing dependencies from both past and future contexts is essential for making accurate predictions. The architecture of BiLSTM used in this study is similar to that of LSTM in Fig \ref{lstm}, with the only addition being a   `Bidirectional' wrapper layer around each LSTM cell. 

Both the LSTM and BiLSTM models mentioned in Section \ref{lstm section} and Section \ref{bilstm section} were trained using 50 epochs with a batch size of 256. The architecture consisted of LSTM/BiLSTM 2 layers with 11 units, activated by ReLU, followed by a Dense layer with 64 units. The model was optimized using Adam and trained to minimize MSE, with MAE serving as a metric. Checkpointing and early stopping were implemented for performance monitoring.

\subsection{LSTM + CNN model}\label{lstm hyb}
\begin{figure}[h]
    \centering
    \captionsetup{justification=centering}
    \includegraphics[width=0.4\textwidth]{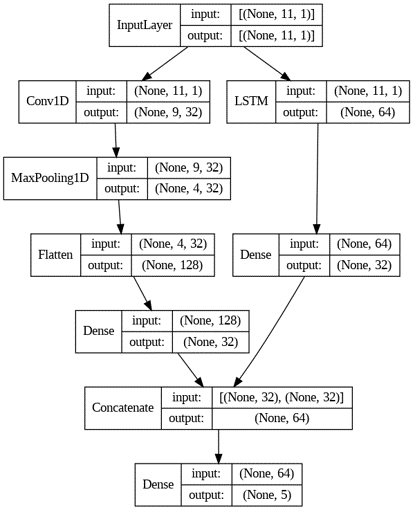}
    \caption{LSTM + CNN architecture}
    \label{lstm+cnn}
\end{figure}

The model architecture illustrated by Fig. \ref{lstm+cnn} integrates both Convolutional Neural Network (CNN) and Long Short-Term Memory (LSTM) components to leverage their respective strengths in capturing spatial and temporal patterns from input time-series data. The CNN segment initiates by convolving over the input data, extracting relevant spatial features through one-dimensional convolutional operations, followed by max-pooling and flattening layers to distill significant patterns along the temporal dimension. These spatial features are then further refined through a densely connected layer, augmenting their representational capacity. Simultaneously, the LSTM component processes the same input data, specializing in capturing temporal dependencies across sequential observations. Comprising two LSTM layers, this segment adeptly captures bidirectional sequential information, with the first layer capturing information from both past and future directions and the second layer consolidating this bidirectional information into a concise representation. Subsequently, another densely connected layer refines the extracted features from the LSTM pathway. Finally, the outputs from both the CNN and LSTM pathways are concatenated, merging spatial and temporal insights into a unified feature space. This concatenated feature vector encapsulates a comprehensive representation of the input sequence, empowering the model to discern intricate spatial and temporal patterns crucial for accurate prediction or classification tasks.\par

The CNN-LSTM hybrid model was chosen for flight delay prediction due to its ability to capture both spatial and temporal patterns inherent in the flight data. Flight delays are influenced by spatial factors such as airport locations and weather conditions, which the CNN component adeptly extracts through convolutional operations on the input time-series data. Additionally, the LSTM component excels at capturing temporal dependencies and sequential patterns, essential for comprehending recurring delays or disruptions in flight schedules. By integrating both spatial and temporal insights into a unified feature space through concatenation, the model gains a holistic understanding of flight data, enabling it to discern and leverage complex spatial and temporal relationships for accurate flight delay prediction. This integrated approach enhances predictive performance compared to traditional machine learning models, making it well-suited for flight delay prediction tasks.

In training the LSTM-CNN mentioned in Section \ref{lstm hyb}, we utilized 50 epochs with a batch size of 256.  Checkpointing and early stopping were employed for performance monitoring. Evaluation of the test set revealed the models' efficacy in predicting flight delays, with MAE providing insights into prediction across flight attributes. 

\section{RESULTS and Discussion}

\subsection{Evaluation Metric}
In evaluating the performance of our models, we utilized the Mean Absolute Error (MAE) metric, which quantifies the average magnitude of errors between predicted and actual values. Mathematically, MAE is expressed as:

\[
MAE = \frac{1}{n} \sum_{i=1}^{n} |y_{\text{pred}, i} - y_{\text{true}, i}|
\]

where \( y_{\text{pred}, i} \) represents the predicted value, \( y_{\text{true}, i} \) denotes the true value, and \( n \) is the number of samples.

The MAE provides a straightforward measure of the model's predictive accuracy, with lower values indicating better performance. We employed this metric to assess the efficacy of our models in accurately predicting flight delays and it is easy to interpret results using minutes.

Mean Squared Error (MSE) was employed as the loss function in our model. MSE quantifies the average squared magnitude of errors between predicted ($\hat{y}$) and actual ($y$) values, as represented by the equation:

\[
MSE = \frac{1}{n} \sum_{i=1}^{n} (\hat{y}_i - y_i)^2
\]

This metric is particularly useful for training models due to its property of giving higher weighting to lower frequency values. In scenarios where there are numerous occurrences of zero values, MSE effectively accounts for these instances and provides a comprehensive measure of the model's performance.

\subsection{Overall Arrival Delay}
\begin{figure}[htp]
    \centering
    \captionsetup{justification=centering}
    \includegraphics[width=9cm]{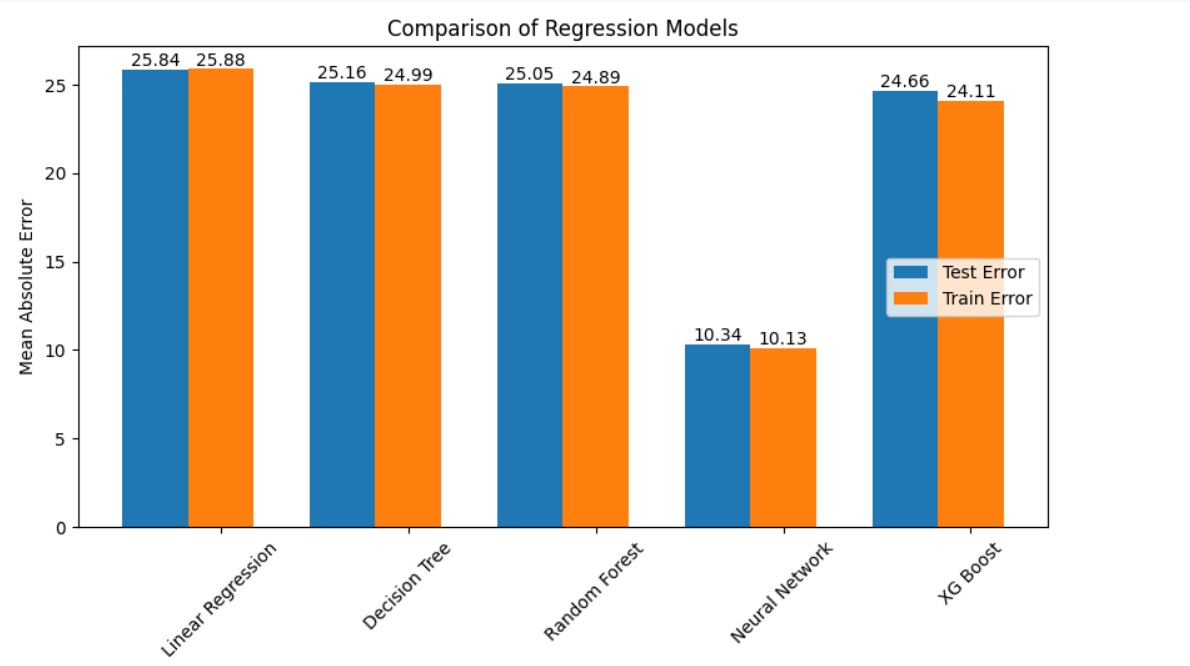}
    \caption{Error Predicting Overall Arrival Delay}
    \label{fig:overall}
\end{figure}

As shown in Fig.\ref{fig:overall}, the Neural Network performed best. However, none of the models provided useful delay predictions.

\subsection{Analysis of Individual Delay Components}
\begin{table}[ht]
\centering
\small
\caption{Comparison of Total Model Errors}
\begin{tabular}{@{}lcc@{}}
\toprule
Model & MSE & MAE \\
\midrule
LSTM & \textbf{361.833} & 10.022 \\
Bidirectional LSTM & 365.645 & 10.296 \\
LSTM + CNN Hybrid & 367.868 & \textbf{9.684} \\
Neural Network* & 371.474 & 10.340 \\
\bottomrule
\end{tabular}
\end{table}

*Best Non-Time-Series Model \\

In evaluating individual delay components, we calculated the average errors across them. Although the LSTM + CNN hybrid architecture showcased superior performance when compared using the MAE as a metric, the LSTM model was prioritized due to its lower Mean Squared Error (MSE). This is due to the nature of our dataset, which includes numerous values close to zero. The MSE penalizes deviations more heavily, thereby not favoring models that guess close to zero for most inputs.

\begin{table}[ht]
\centering
\caption{LSTM Test Results for Individual Delay Components}
\begin{tabular}{lccc}
\toprule
Delay Component & True Mean & Mean of Predictions & MAE \\
\midrule
Carrier & 15.877 & 16.288 & 17.050 \\
Weather & 1.978 & 2.246 & 3.979 \\
Security & 0.130 & 0.141 & 0.270 \\
NAS & 10.914 & 11.512 & 9.475 \\
Late Aircraft & 19.374 & 18.031 & 19.338 \\
\bottomrule
\end{tabular}
\label{tab:LSTM Perf}
\end{table}

Table.\ref{tab:LSTM Perf} presents the LSTM model's performance on individual delay components. Despite high MAE values, the proximity between the true mean and the mean of predictions suggests the reasonable ability of the model to predict the delay. Unfortunately, the high MAE values suggest very low predictive power of even our best models. We believe this high error is influenced by the distribution of delay durations (for example the true mean of delay due to security is only 0.130 whereas the true mean of delay due to late aircraft is 19.374) across components.

\section{Conclusion and future scope}

While our time-series approaches demonstrated slightly improved performance compared to baseline regression models, it is imperative to acknowledge their limitations in providing meaningful predictions. Several factors may contribute to the observed challenges.

Firstly, the unprecedented impact of COVID-19 on the data could significantly influence the performance of our models. The abrupt disruptions and fluctuations caused by the pandemic may have introduced unforeseen complexities, making it challenging for the models to capture underlying patterns accurately. Secondly, the presence of outlier values within the dataset poses another potential obstacle. Outliers can distort the learning process and adversely affect the predictive capabilities of the models, thereby undermining the reliability of our predictions. While we removed outliers based on total arrival delay, we did not consider outliers for each individual delay component. Moreover, our approach of regression on delay components introduces additional intricacies compared to prior classification-based methodologies. While this granularity offers insights into specific aspects of delays, it also amplifies the complexity of the prediction task, potentially leading to less actionable results.

To address these challenges and enhance the effectiveness of our predictive models, several avenues for future research and development can be explored.

Firstly, incorporating specialized techniques for handling COVID-19 data is paramount. Strategies such as feature engineering to capture pandemic-related trends or integrating external data sources containing relevant information could bolster the resilience of our models against the disruptive effects of such events. Furthermore, exploring different outlier detection and removal methods can improve the robustness of our models. Adopting techniques tailored to the characteristics of our dataset, such as robust statistical measures or anomaly detection algorithms, can help mitigate the adverse effects of outliers on the model performance. In addition, reassessing the granularity of our prediction task warrants consideration. While regression on delay components provides detailed insights, simplifying the task by focusing on total delay may offer a more tractable approach, particularly if the intricacies of component-level predictions prove to be overly challenging or resource-intensive.

In conclusion, by addressing the aforementioned issues and embracing these future directions, we can refine our predictive models and empower them to deliver more accurate and actionable insights, thereby enhancing their utility in real-world applications.


\appendix
\clearpage 

\section{Dataset Details}

\begin{table}[htp]
 
    \captionsetup{justification=centering}
    \centering
    \caption{Dataset Attributes}
    \begin{tabular}{|p{4cm}|p{4cm}|}
    \hline
    \textbf{Name} & \textbf{Description} \\
    \hline
    FL\_DATE & Flight Date (yyyymmdd) \\
    \hline
    AIRLINE\_CODE & Unique Carrier Code \\
    \hline
    DOT\_CODE & An identification number assigned by US DOT to identify a unique airline (carrier) \\
    \hline
    FL\_NUMBER & Flight Number \\
    \hline
    ORIGIN & Origin Airport \\
    \hline
    ORIGIN\_CITY & Origin Airport, City Name \\
    \hline
    DEST & Destination Airport \\
    \hline
    DEST\_CITY & Destination Airport, City Name \\
    \hline
    CRS\_DEP\_TIME & CRS Departure Time (local time: hhmm) \\
    \hline
    DEP\_TIME & Actual Departure Time (local time: hhmm) \\
    \hline
    DEP\_DELAY & Difference in minutes between scheduled and actual departure time \\
    \hline
    TAXI\_OUT & Taxi Out Time, in Minutes \\
    \hline
    WHEELS\_OFF & Wheels Off Time (local time: hhmm) \\
    \hline
    WHEELS\_ON & Wheels On Time (local time: hhmm) \\
    \hline
    TAXI\_IN & Taxi In Time, in Minutes \\
    \hline
    CRS\_ARR\_TIME & CRS Arrival Time (local time: hhmm) \\
    \hline
    ARR\_TIME & Actual Arrival Time (local time: hhmm) \\
    \hline
    ARR\_DELAY & Difference in minutes between scheduled and actual arrival time \\
    \hline
    CANCELLED & Cancelled Flight Indicator (1=Yes) \\
    \hline
    CANCELLATION\_CODE & Specifies The Reason For Cancellation \\
    \hline
    DIVERTED & Diverted Flight Indicator (1=Yes) \\
    \hline
    CRS\_ELAPSED\_TIME & CRS Elapsed Time of Flight, in Minutes \\
    \hline
    ELAPSED\_TIME & Elapsed Time of Flight, in Minutes \\
    \hline
    AIR\_TIME & Flight Time, in Minutes \\
    \hline
    DISTANCE & Distance between airports (miles) \\
    \hline
    DELAY\_DUE\_CARRIER & Carrier Delay, in Minutes \\
    \hline
    DELAY\_DUE\_WEATHER & Weather Delay, in Minutes \\
    \hline
    DELAY\_DUE\_NAS & National Air System Delay, in Minutes \\
    \hline
    DELAY\_DUE\_SECURITY & Security Delay, in Minutes \\
    \hline
    DELAY\_DUE\_LATE\_AIRCRAFT & Late Aircraft Delay, in Minutes \\
    \hline
    \end{tabular}

    \label{tab:dataset_attributes}
\end{table}

\section{Models}
\subsection{Random Forest Regressor}

\begin{figure}[h]
    \centering
    \captionsetup{justification=centering}
    \includegraphics[width=0.4\textwidth]{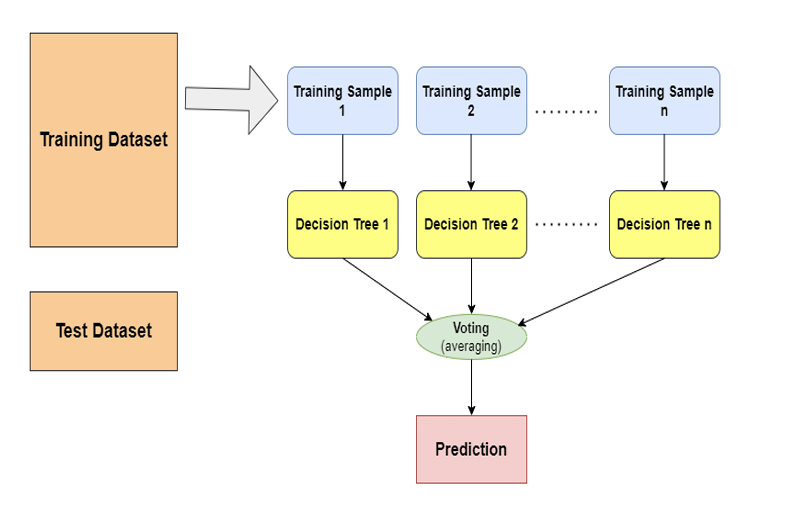}
    \caption{Random Forest Model \cite{15}}
    \label{fig:RF}
\end{figure}

\subsection{XGBoost Regressor}
\begin{figure}[h]
    \centering
    \captionsetup{justification=centering}
    \includegraphics[width=0.45\textwidth]{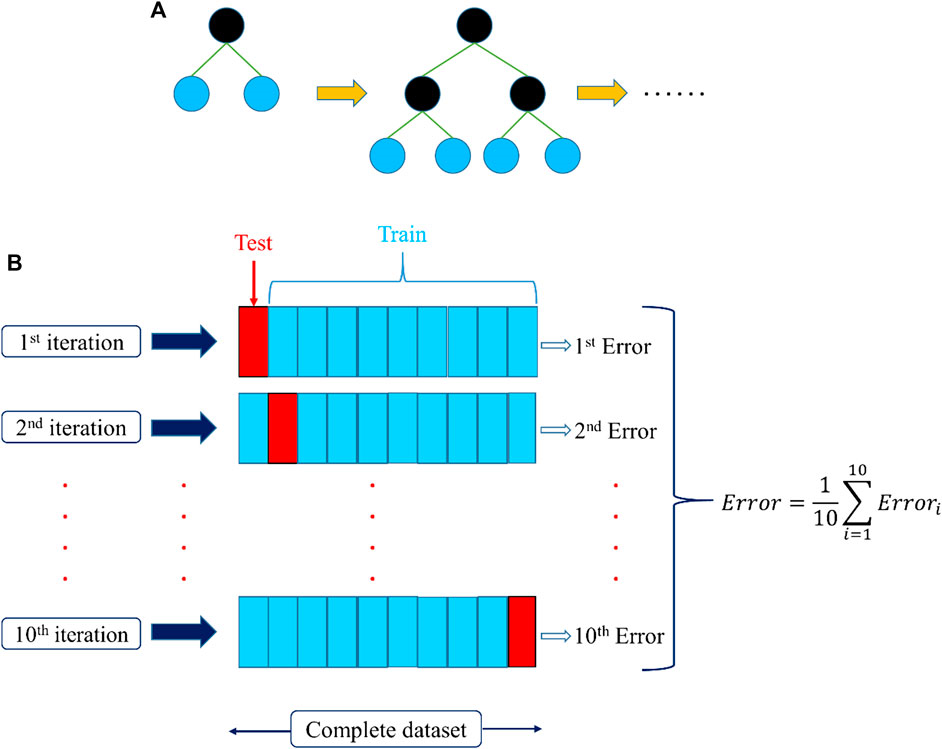}
    \caption{XGBoost Regressor Model \cite{14}}
    \label{fig:xgboost}
\end{figure}

\begin{equation}
y_{-ki} = y_{-(k-1)i} + \eta f_k \label{eq:estimator}
\end{equation}

\begin{equation}
f_{obj} = \gamma Z + \sum_{a=1}^{Z} [g_a \omega_a + \frac{1}{2} (h_a + \lambda) \omega_a^2] \label{eq:fk}
\end{equation}

\begin{equation}
G = \frac{1}{2} \left[ \frac{O^2L}{P_L} + \lambda + \frac{O^2R}{P_R} + \lambda + \left( \frac{OL + OR}{P_L + P_R} \right)^2 P_L + P_R + \lambda \right] \label{eq:gain}
\end{equation}
These are the equations involved in the calculations of the XGBoost Regressor model.
Equation \ref{eq:estimator} represents the estimation of the output ($y_{-ki}$) by the XGBoost model at the $i$th data point in the $k$th stage of training. Here, $y_{-ki}$ is the predicted output at the $i$th data point after the $k$th stage of training, with $y_{-(k-1)i}$ representing the predicted output after the $(k-1)$th stage. The learning rate $\eta$ controls the contribution of each tree to the final prediction, while $f_k$ denotes the function representing the $k$th tree added to the model. In Equation \ref{eq:fk}, the objective function ($f_{obj}$) used in the XGBoost algorithm is defined. It comprises the sum of the loss function and a regularization term essential for optimizing the model during training. The complexity parameter $\gamma$ and regularization parameter $\lambda$ are included to control the model's complexity and amount of regularization applied. Additionally, the gain parameter ($G$) in Equation \ref{eq:gain} calculates the quality of a split during tree building based on the improvement in impurity or loss function. This equation incorporates various parameters such as the sum of squared gradients ($O^2L$ and $O^2R$) and the number of data points in the left and right child nodes ($P_L$ and $P_R$), along with the regularization parameter $\lambda$.

\subsection{Aritificial Neural Network}

\begin{figure}[h]
    \centering
    \captionsetup{justification=centering}
    \includegraphics[width=0.5\textwidth]{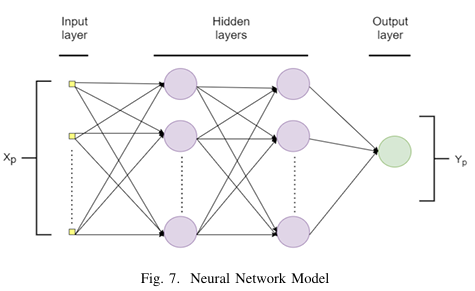}
    \caption{Aritificial Neural Network Architecture \cite{15}}
    \label{fig:ann}
\end{figure}

\end{document}